\documentclass[sensors,article,submit,moreauthors,pdftex]{Definitions/mdpi} 

\usepackage{graphics} 
\usepackage{epsfig} 
\usepackage{mathptmx} 
\usepackage{times} 
\usepackage{amsmath} 
\usepackage{amssymb}  
\usepackage{textcomp}
\usepackage{multirow}
\usepackage{colortbl}
\usepackage{xcolor}
\usepackage{hhline}
\usepackage{bm}
\usepackage{hyperref}
\usepackage{lipsum}
\usepackage{mathtools}
\usepackage{cuted}
\usepackage{subfig}
\usepackage{comment}

\usepackage[colorinlistoftodos, textwidth=3.7cm]{todonotes}
\reversemarginpar

\graphicspath{{figs/}{evaluation/}}

\preto{\abstractkeywords}{\nolinenumbers}

\renewcommand{\vec}[1]{\mathbf{#1}}
\newcommand{\newcolor}[1]{\color{black}}
\definecolor{red}{cmyk}{0,0.89,0.94,0.28}

\firstpage{1} 
\pubvolume{1}
\issuenum{1}
\articlenumber{0}
\pubyear{2021}
\copyrightyear{2021}
\externaleditor{Academic Editor: Firstname Lastname} 
\datereceived{} 
\dateaccepted{} 
\datepublished{} 
\hreflink{https://doi.org/} 

\Title{OctoPath: An OcTree Based Self-Supervised Learning Approach to Local Trajectory Planning for Mobile Robots}

\TitleCitation{Title}

\Author{Bogdan~Tr\u{a}snea$^{1,2,*}$, Cosmin~Gineric\u{a}$^{1,2}$, Mihai Zaha$^{1,2}$, Gigel~M\u{a}ce\c{s}anu$^{1,2}$, Claudiu~Pozna$^{1}$ and Sorin~Grigorescu$^{1,2}$}

\AuthorNames{Bogdan~Tr\u{a}snea, Cosmin~Gineric\u{a}, Mihai Zaha, Gigel~M\u{a}ce\c{s}anu, Claudiu~Pozna and Sorin~Grigorescu}

\address{%
$^{1}$ \quad  Robotics, Vision and Control Laboratory (ROVIS) at Transilvania University of Brasov; bogdan.trasnea@unitbv.ro, gigel.macesanu@unitbv.ro, cp@unitbv.ro, s.grigorescu@unitbv.ro\\
$^{2}$ \quad Elektrobit Automotive; bogdan.trasnea@elektrobit.com, sorin.grigorescu@elektrobit.com}

\corres{Correspondence: bogdan.trasnea@unitbv.ro}

\abstract{Autonomous mobile robots are usually faced with challenging situations when driving in complex environments. Namely, they have to recognize the static and dynamic obstacles, plan the driving path and execute their motion. For addressing the issue of perception and path planning, in this paper, we introduce \textit{OctoPath}, which is an encoder-decoder deep neural network, trained in a self-supervised manner to predict the local optimal trajectory for the ego-vehicle. Using the discretization provided by a 3D octree environment model, our approach reformulates trajectory prediction as a classification problem with a configurable resolution. During training, OctoPath minimizes the error between the predicted and the manually driven trajectories in a given training dataset. This allows us to avoid the pitfall of regression-based trajectory estimation, in which there is an infinite state space for the output trajectory points. Environment sensing is performed using a 40-channel mechanical LiDAR sensor, fused with an inertial measurement unit and wheels odometry for state estimation. The experiments are performed both in simulation and real-life, using our own developed GridSim simulator and RovisLab's Autonomous Mobile Test Unit platform. We evaluate the predictions of OctoPath in different driving scenarios, both indoor and outdoor, while benchmarking our system against a baseline hybrid A-Star algorithm and a regression-based supervised learning method, as well as against a CNN learning-based optimal path planning method.}


\begin{document}


\section{Introduction}
\label{sec:introduction}
 
Recent developments in the fields of deep learning and artificial intelligence have aided the autonomous driving domain's rapid advancement. Autonomous vehicles (AVs) are robotic systems that can navigate without the need for human intervention. The deployment of AVs is predicted to have a major impact on the future of mobility, bringing a variety of benefits to daily life, such as making driving simpler, increasing road network capacity, and minimizing vehicle-related crashes.

For Advanced Driver Assistance Systems(ADAS) systems and autonomous robot control, one of the top priorities is ensuring functional safety. When a car is driving, it encounters varieties of dynamic traffic scenarios in which the moving objects in the environment may pose a risk to safe driving. The car must consider all threats present in the surrounding environment in order to create a collision-free path and determine the next steps based on it. Due to the complexity of such a task, deep learning models have been used to aid in solving it. There are several conceptually different self-driving architectures, namely end2end learning~\cite{xu2017end}, Deep Reinforcement Learning~\cite{Jaritz_ICRA2018} (DRL) and the sense-plan-act pipeline~\cite{pendleton2017perception}.

\begin{figure*}
	\centering
	\begin{center}
		\includegraphics[scale=0.96]{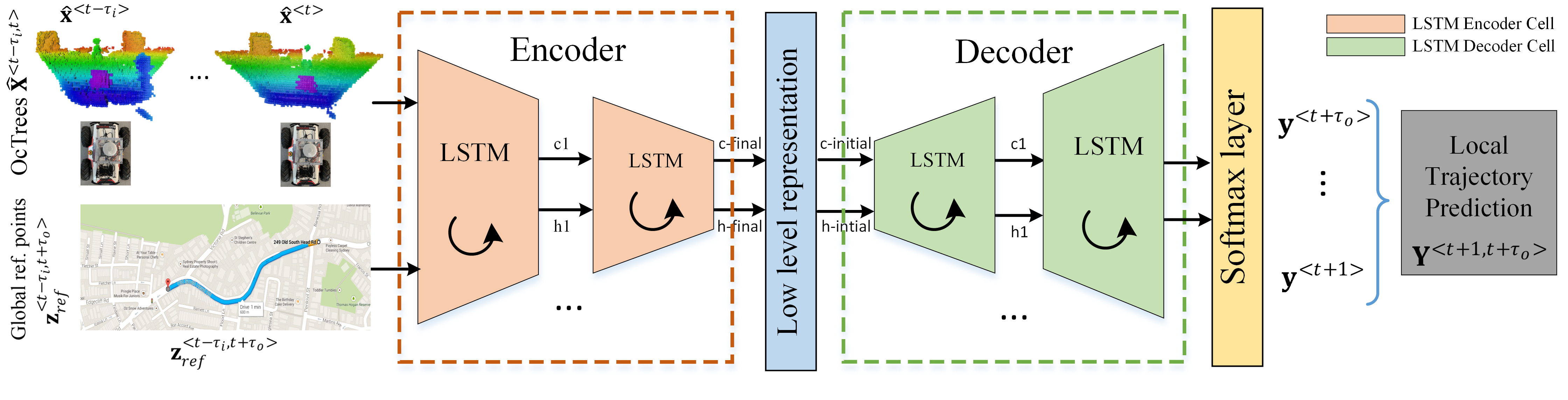}
		\caption{\textbf{Local trajectory prediction using a neural network encoder-decoder architecture.} The training data consists of sequences of OcTrees $({\vec{X}}^{<t-\tau_i, t>})$ and points from the reference path $(\vec{Z}_{ref}^{<t-\tau_i, t+\tau_o>})$. Future trajectory points $(\vec{Y}^{<t+1, t+\tau_o>})$ are used to calculate the labels in a self-supervised manner. The encoder-decoder network architecture is designed to take advantage of the neural network's ability to learn effective temporal representations. The input sequences are passed through an encoder network, which maps raw inputs to a hidden feature representation known as a \textit{thought vector}, and a decoder network, which takes this feature representation as input, processes it and outputs a trajectory prediction.}
        \label{fig:neural_network_diagram}
	\end{center}
\end{figure*}

In the sense-plan-act case, an important component is the one that plans the future driving path of the ego-vehicle. Determining a safe path over a finite prediction horizon is a key aspect when employing a control strategy, especially when considering dynamic and static obstacle avoidance. The main problem is split into smaller sub-problems in a modular pipeline, with each module intended to solve a particular task and provide the result as input to the next component. Therefore, AVs must have the ability to sense their surroundings and form an adequate environment model which precisely represents the dynamic and stationary objects. Afterward, it needs to plan its path, which is defined as the AV's ability to find a collision-free route between the current position and the desired destination. Finally, it needs to act based on the computed path, by applying the appropriate control signals (acceleration and steering) for the AV.

In this paper, we address the path planning component from the perception-planning-action pipeline. Fig.~\ref{fig:neural_network_diagram} highlights the block diagram of the proposed concept. The currently proposed method, coined \textit{OctoPath}, is self-supervised and aims to combine the configurable discretization of an octree-based environment model with a classification-based encoder-decoder RNN architecture. It takes as input a sequence of sensor measurements, together with the current segment of a reference trajectory, building upon the RNN encoder-decoder architecture which has shown excellent performance for sequence-to-sequence tasks. 

As opposed to~\cite{Grigorescu_RAL_2019}, we now sense the world in 3D using an octree representation, and we no longer use convolutional layers for processing the input sequences, as this intermediate representation has been taken over by the fixed state vector between the encoder and the decoder of our architecture. 

The key contributions of this paper are:

\begin{enumerate}
    \item based on the octree environment model, we provide a solution for estimating local driving trajectories by reformulating the estimation task as a classification problem with a configurable resolution;
    \item we define an encoder-decoder deep neural network topology for predicting desired future trajectories, which are obtained in a self-supervised fashion;
    \item we leverage the innate property of the state vector between the encoder and the decoder to represent a learned sequence of trajectory points constrained by road topology.
\end{enumerate}

The rest of this paper is organized as follows. Section~\ref{sec:related_work} covers the related work. In Section~\ref{sec:method}, the trajectory estimation problem for mobile robots is briefly described (\ref{sec:problem_definition}), together with the octree environment model (\ref{sec:octree_env_model}) and the RovisLab's AMTU kinematic model (\ref{sec:kinematics}). Our choice of recurrent encoder-decoder neural network architecture(\ref{sec:rnn_encoder_decoder_architecture}) is detailed in Section~\ref{sec:octopath}, in combination with the training setup (\ref{sec:training}) using the gathered mobile robot data. Performance evaluation can be found in Section~\ref{sec:results}, where we first overview the experimental setup (\ref{sec:experiments_overview}) and detail the related experiments (\ref{sec:experiments_I}, \ref{sec:experiments_II}) , and second, we analyze the deployment of our OctoPath network on the NVidia AGX Xavier (\ref{sec:deployment}) and an ablation study of varying the octree resolution and the network model parameters (\ref{sec:ablation}). Sections~\ref{sec:conclusions} covers the conclusions of the paper.

\section{Related Work}
\label{sec:related_work}

Deep learning~\cite{bengio2013representation, hessel2018rainbow} has emerged as the driving force behind many of the new developments in recent years, with notable advancements in computer vision, robotics, and natural language processing. DNNs have effectively learned representations that generalize well for different scenarios arising from real data when applied to a variety of machine learning tasks.

Among the different kinds of DNN architectures available, recurrent neural networks (RNN) are used to analyze the structure of time series data, such as text, or video streams~\cite{cho2014learning}. All of these advancements in the field of deep learning have obviously influenced the growth of intelligent vehicles. The first automated cars were described in the late 1980s and early 1990s, and following these first implementations, various control architectures for automated driving were proposed~\cite{gonzalez2015review}. The driving functions are generally applied as perception-planning-action pipelines, as seen in Grigorescu et. al's deep learning for autonomous driving survey~\cite{grigorescu2020survey}, but recent approaches based on end2end learning have also been proposed, though most as research prototypes.

In this context, end2end learning is defined as developing and training a complex neural network to directly map input sensory data to vehicle commands~\cite{amini2020learning}. The authors of~\cite{pan2020imitation} present an end-to-end imitation learning system for off-road autonomous driving by using only low-cost onboard sensors, having their DNN policy trained for agile driving on a predefined obstacle-free track. Since self-driving cars must manage roads with complex barriers and unclear lane borders, this strategy restricts the applicability of their system to autonomous driving. In the PilotNet algorithm proposed by Bojarski et al.~\cite{bojarski2016end}, the input images are directly mapped to the vehicle's steering control.

Because large annotated datasets are needed to train such deep networks, the alternative of self-supervised learning approaches has also been employed. In~\cite{kahn2020badgr}, BADGR - Berkeley Autonomous Driving Ground Robot, an End2End self-supervised learning system, was created to navigate in real-world situations with geometrically distracting obstacles (such as tall grass). It can also take into account terrain preferences, generalize to new environments, and improve on its own by collecting more data.

DRL (Deep Reinforcement Learning) is an algorithm or a system in which agents learn to act by communicating with their surroundings. Although DRL does not use training data, it maximizes a cumulative reward that quantifies its behavior \cite{mnih2015human}. In \cite{panov2018grid}, well-established route planning techniques are correlated with novel neural learning approaches to find the best path to a target location within a square grid. A combination of global guidance and a local RL-based planner is presented in~\cite{wang2020mobile}. A major downside of DRL is that such an architecture is difficult to train in real-life scenarios due to the interaction constraint and tends to generalize on specific driving scenarios (e.g. highway driving). Furthermore, because the input from the sensors is straightly mapped to actuators, these systems' functional safety is normally difficult to monitor~\cite{salay2018analysis}.

Path planning in mobile robotics has been a topic of research for decades~\cite{steffi2021robot}, and it is categorized into global and local planning based on the scope and executability of the plan~\cite{cai2020mobile}. The survey in~\cite{pendleton2017perception} offers a broad overview of route planning in the automotive industry. It focuses on the task planner, behavior planner, and motion planner, which are all taxonomy elements of path planning. It does not, however, provide a study of deep learning technology, despite increased interest in using deep learning technologies for route planning and behaviour arbitration in the state-of-the-art literature.

Novel DNN-based path planning methods which have been developed are based on biologically inspired cognitive architectures~\cite{li2019survey}, with the three primary methods being swarm intelligence, evolutionary algorithms, and neurodynamics. In the case of~\cite{wang2020neural}, an optimal path planning algorithm based on convolutional neural networks (CNN) and random-exporing trees (RRT) is presented. Their approach, called Neural RRT*, is a framework for generating the sampling distribution of the optimal path under several constraints. In our previous work on local state trajectory estimation~\cite{Grigorescu_RAL_2019}, we used a multi-objective neuro-evolutionary approach to train a regression-based hybrid CNN-LSTM architecture using sequences of 2D occupancy grids.

Since motion planning can also be viewed as a sequence to sequence mapping problem, or as a sequence generation task, RNNs have been proposed for modeling the driving trajectories~\cite{wu2017modeling, park2018sequence, ma2019trafficpredict}. The input sequence in this technique is composed of the most recent sensor readings, while the output sequence contains the future trajectory points. In contrast to traditional neural networks, RNN memory cells have a time-dependent feedback loop. In order to use RNNs for predicting a future trajectory, each separate point is considered a state, which further implies that the whole trajectory is represented as a sequence. The transition from one state to another is strictly constrained by the topology of the network~\cite{altche2017lstm}.

Many approaches address the task of predicting the trajectory of vehicles surrounding the ego-car: in~\cite{deo2018multi}, the authors suggested a multi-modal trajectory prediction system for surrounding vehicles, which assigns confidence values to vehicle maneuvers and generates a multi-modal distribution over future motion based on those values. The authors of~\cite{kim2017probabilistic} suggest an LSTM that predicts the location of cars in an occupancy grid at discrete intervals of $0.5s$, $1s$ and $2s$ in the future, while Park et al.~\cite{park2018sequence} employs an encoder-decoder architecture to generate the $K$ most likely trajectory candidates over an occupancy grid map using the beam search method.

Most of the RNN solutions proposed for solving the task of trajectory estimation need a discrete environment model. In this work, the proposed environment model is based on octrees~\cite{hornung2013octomap, han2018towards} and uses probabilistic occupancy estimation. The main advantages of using this model are that it explicitly represents not only occupied space but also free and unknown areas and that it enables a compact memory representation and configurable resolutions. In~\cite{vanneste20143dvfh}, the authors used an octree-based model to determine the surrounding obstacle locations in real-time, and use it for path planning and robot motion generation. The authors of~\cite{zhang2020multi} present the advantages of having an environment with multiple resolutions and a uniform octree representation mechanism of models from various sensors.

\section{Method}
\label{sec:method}

\begin{figure}
	\centering
	\begin{center}
		\includegraphics[scale=0.71]{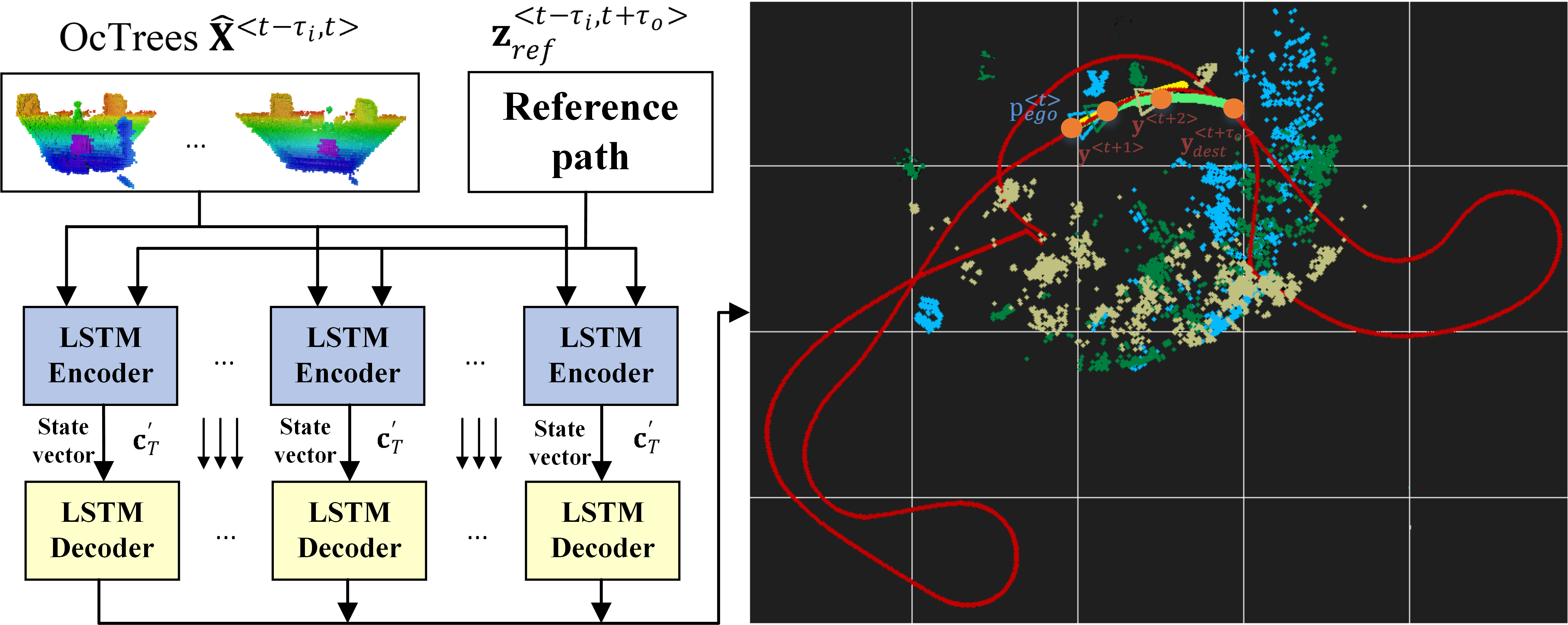}
		\caption{\textbf{Estimating local trajectories for autonomous robots and vehicles using encoder-decoder recurrent neural networks.} Considering the current ego vehicle's position $\vec{p}_{ego}^{<t>}$, an input sequence of octrees $\vec{X}^{<t-\tau_i, t>}=[\vec{x}^{<t-\tau_i>}, ..., \vec{x}^{<t>}]$, the current sequence from the reference route $\vec{z}_{ref}^{<t-\tau_i, t+\tau_o>}$ and a desired destination from the reference path $\vec{y}_{dest}^{<t+\tau_o>}$, the aim is to estimate a driving trajectory $\vec{Y}^{<t+1, t+\tau_o>}=[\vec{y}^{<t+1>}, ..., \vec{y}_{dest}^{<t+\tau_o>}]$.}
        \label{fig:problem_description}
	\end{center}
\end{figure} 

A variable's value is defined either as a discrete sequence defined in the $<t, t+k>$ time interval, where $k$ represents the length of the sequence, or as a single discrete time step $t$, written as superscript $<t>$. The value of a trajectory output variable $\vec{y}$, for example, may be specified at discrete time $t$ as $\vec{y}^{<t>}$ or within a sequence interval $\vec{Y}^{<t, t+k>}$.

\subsection{Problem Definition: Local Trajectory Prediction}
\label{sec:problem_definition}

A trajectory is defined as a time-and-velocity-parameterized sequence of states visited by the vehicle. Local trajectory planning (also known as local trajectory generation) is concerned with the real-time planning of a vehicle's transition from its current state to the next while avoiding obstacles and meeting the vehicle's kinematic limitations, over a prediction horizon. Depending on the speed and line-of-sight of the vehicle's on-board sensors, the route planner module produces an estimated optimal trajectory from the vehicle's current location, with a look-ahead distance, during each planning cycle.

Figure~\ref{fig:problem_description} depicts an illustration of the local state trajectory prediction task for autonomous driving. The task is to learn a local trajectory for navigating the ego-vehicle to destination coordinates $\vec{y}_{dest}^{<t+\tau_o>}$ given a sequence of octrees $\vec{x}^{<t-\tau_i>}: \mathbb{R}^3 \times \tau_i$, the current sequence from the reference route $\vec{z}_{ref}^{<t-\tau_i, t+\tau_o>}: \mathbb{R}^2 \times \tau_i$, the position of the ego-vehicle $\vec{p}_{ego}^{<t>} \in \mathbb{R}^2$ in $\vec{x}^{<t>}$, and the destination coordinates $\vec{p}_{dest}^{<t>} \in \mathbb{R}^2$ at time $t$. The length of the input data sequence is $\tau_i$, and $\tau_o$ is the number of time steps for which the ego-vehicle's trajectory is computed. The reference path is represented by a collection of points in the global coordinate system, which depict the route that the robot has to follow. An example of such a generated path using our vehicle mission planner tool can be found in the top-left side of Fig.~\ref{fig:outdoor_testing}.

The local trajectory problem can be expressed as a classification problem with N classes, where N is determined by the OcTree resolution (density of points in the environment space) and the prediction horizon (how far away is the destination point). It is a multi-class classification problem in which each time step's point on the trajectory is selected sequentially from an input sequence of octree environment snapshots and points from the reference route.

In other words, we pursue a desired local navigation trajectory of the ego-vehicle from any arbitrary starting point $\vec{p}_0^{<t>}$ (which is a coordinate in the current input octree) to $\vec{y}_{dest}^{<t+\tau_o>}$, with the following properties:

\begin{itemize}
	\item the longitudinal velocity $v^{<t, t+\tau_o>}$ is maximal and is contained within the bounds $[v_{min}, v_{max}]$;
	\item the total distance between consecutive trajectory points is minimal: $||(p_0^{<t>} - \vec{y}_{dest}^{<t>}) + \sum\limits_{i=0}^{\tau_0} (\vec{y}_{dest}^{<t+i>} - \vec{y}_{dest}^{<t+i+1>}) ||$;
	\item the lateral velocity $v_{\delta}^{<t, t+\tau_o>}$ is minimal. It is is determined by the rate of change of the steering angle $v_{\delta} \in \left[ \dot{\delta}_{min}, \dot{\delta}_{max} \right]$.
	\end{itemize}

The vehicle is modelled based on the kinematics of a skid-steering wheeled mobile robot, with position state $\vec{p}^{<t>} = (p^{<t>}_x, p^{<t>}_y)$ and no-slip assumptions, which is further detailed in section~\ref{sec:kinematics}.

\subsection{Octree Environment Model}
\label{sec:octree_env_model}

Most robotic applications require an environment model, which must be effective in terms of runtime and memory usage, and include free, occupied, and unmapped zones. Sensor models often have range measurement errors, and there may also be apparently random measurements caused by reflections or dynamic barriers. The underlying uncertainty must be taken into account when creating an accurate model of the environment from such noisy data. Multiple uncertain measurements can then be combined to form a reliable estimation of the environment's true state.

An octree is a hierarchical data structure for 3D spatial subdivision that is most commonly used to partition a given 3D space into eight octants by recursively subdividing it. Every node on an octree is the space of a cubic volume that is commonly referred to as a voxel. Each internal node has exactly eight children, and the octree's \textit{resolution} is determined by the minimum voxel scale. If the inner nodes are retained accordingly, the tree can be cut down at any level to acquire a more coarse subdivision. Octrees prevent one of the key weaknesses in fixed grid systems in robotic mapping, the fact that the surroundings should not previously be known, and that the environmental model includes only the assessed volume.

Octrees prevent one of the key weaknesses in fixed grid systems in robotic mapping, the fact that the mapping surroundings should not previously be known, and the environmental model includes only volume measured.

When referring to a laser range finder, for example, the endpoints of the sensor generate occupied space, while the detected region between the sensor and the endpoint is considered to be free space. The occupied space is mapped from the point cloud data packets at the corresponding distance in space for our input LiDAR data. As a result, we use LiDAR data to generate an octree environment model, which depicts free-space (driving area) and inhabited areas in three dimensions. Figure~\ref{fig:indoor_testing} shows an example of input LiDAR data and afferent octree representations queried for occupied voxels together with the projected 2D environment model.

\begin{figure}
	\centering
	\begin{center}
		\includegraphics[scale=1.26]{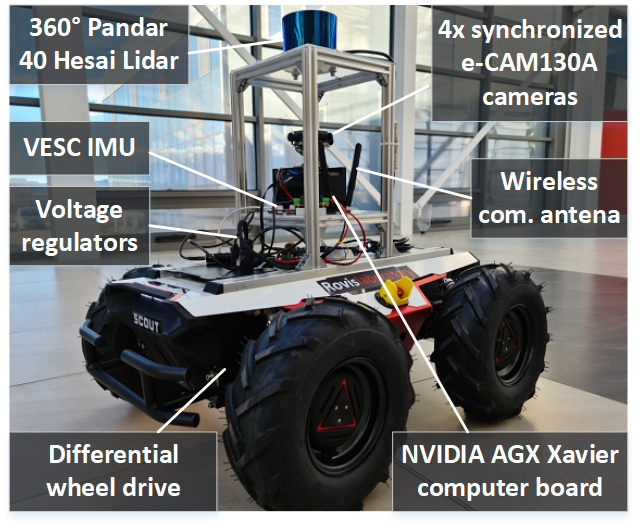}
		\caption{\textbf{RovisLab AMTU (Autonomous Mobile Test Unit).} The robot is a SSWMR (skid-steer wheeled mobile robot) platform equipped with a 360 degree, 40-Channel Hesai Pandar Lidar, 4x e-CAM130A cameras, a Tinkerforge inertial measurement unit (IMU), and an NVIDIA AGX Xavier computer board for real-time data processing and control.}
        \label{fig:rovis_amtu}
	\end{center}
\end{figure}

A central property of our approach is that it allows for efficiency of occupied and free space while keeping the memory consumption low, which is essential for our model car hardware from Fig.~\ref{fig:rovis_amtu}. The octrees have fixed sizes, as required by the neural network input, based on the field of view of the LiDAR sensor. The nodes which are neither occupied nor free (these are always beyond the detected obstacles) are marked as unknown and initialized with zero 0 to prevent them from influencing the inference result. Additionally, we can configure the resolution to a lower value, to reduce the processing times and memory usage even further. Details regarding the impact of varying the octree resolution are provided in~\ref{sec:ablation} and on the right side of Fig.~\ref{fig:ablation_study_octree_resolution}.

\subsection{Kinematics of RovisLab's AMTU as a SSWMR (Skid-steer wheeled mobile robot)}
\label{sec:kinematics}

\begin{figure}
	\centering
	\begin{center}
		\includegraphics[scale=1.41]{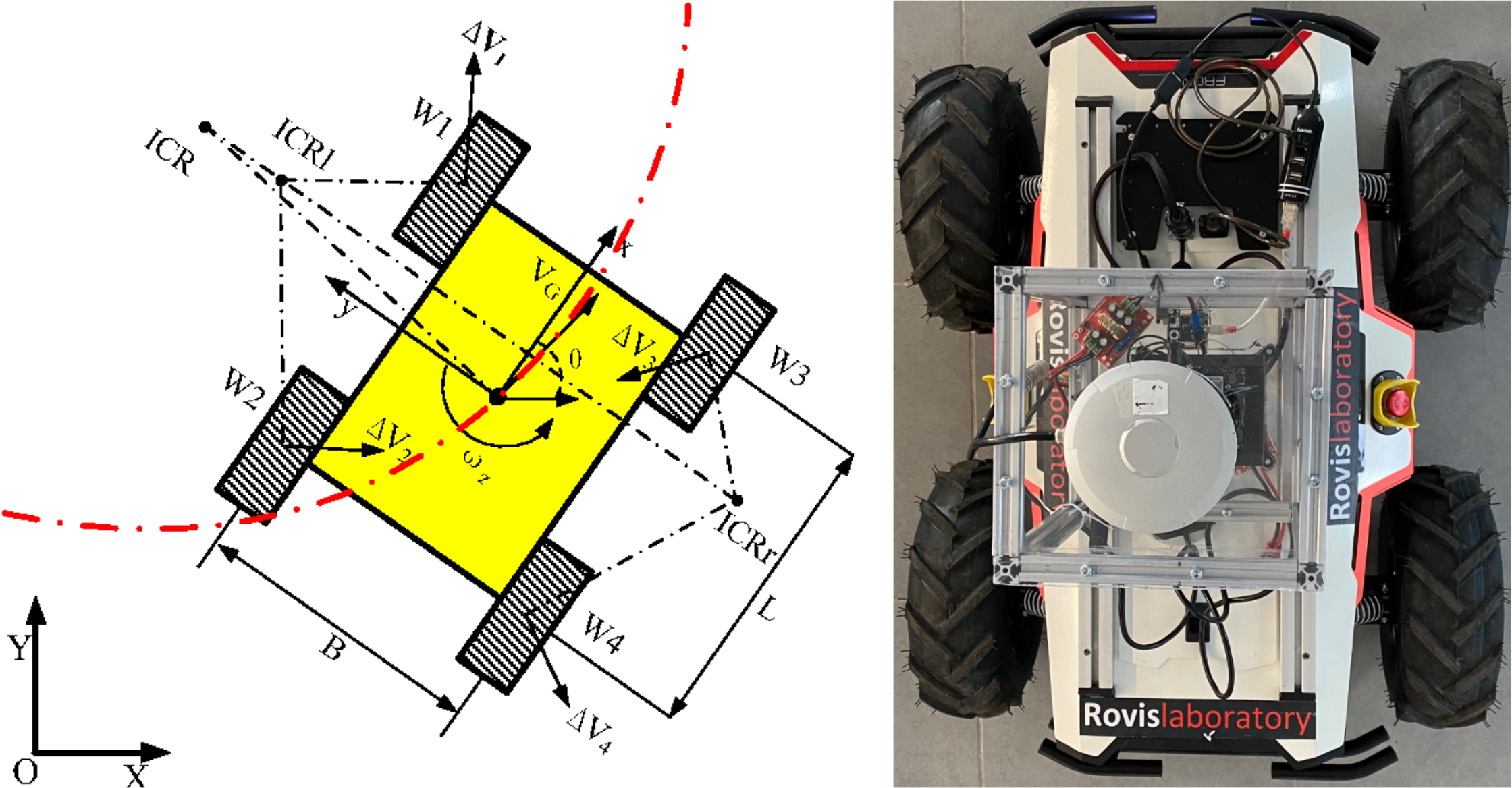}
		\caption{\textbf{A skid-steering mobile robot's kinematics diagram and the top view of RovisLab Autonomous Mobile Test Unit.}}
        \label{fig:skid_steer_kinematic}
	\end{center}
\end{figure}

Figure~\ref{fig:skid_steer_kinematic} shows the schematic diagram of a SSWMR (skid-steer wheeled mobile robot)~\cite{wang2015analysis} together with a top view of RovisLab's AMTU. The following model assumptions are taken into account:

\begin{enumerate}
\item	The robot's mass center is at the geometric center of the body frame;
\item	Each side's two wheels rotate at the same speed;
\item	The robot is operating on a firm ground floor with all four wheels in contact with it at all times.
\end{enumerate}

As shown in Figure~\ref{fig:skid_steer_kinematic}, we define an inertial frame (X, Y) (global frame) and a local (robot body) frame (x, y). Presume the robot movies in a plane with linear velocity $v=(v_{x}, v_{y},0)^{T}$ and rotates with an angular velocity $\omega=(0,0,\omega_{z})^{T}$, both expressed in the local frame. If $q=(X, Y, \theta)^{T}$ is the state vector defining the robot's generalized coordinates (position X and Y, as well as the orientation $\theta$ of the local coordinate frame with respect to the inertial frame), then $\dot{q}=(\dot{X},\dot{Y},\dot{\theta})^{T}$ is the vector of generalized velocities.

The relationship between the robot velocities in both frames is then calculated as follows:

\begin{equation}
\begin{bmatrix}\dot{X} \\ \dot{Y} \\ \dot{\theta} \end{bmatrix} = \begin{bmatrix}\cos\theta & -\sin\theta & 0 \\ \sin\theta & \cos\theta & 0 \\ 0 & 0 & 1 \end{bmatrix} \begin{bmatrix}v_{x} \\ v_{y} \\ \omega_{z} \end{bmatrix}
	\label{eq:robot_vel}
\end{equation}

Because it only specifies free-body kinematics, Equation (\ref{eq:robot_vel}) places no limits on the SSWMR plane movement. As a result, the relationship between wheel velocities and local velocities must be analyzed. For simplicity, the thickness of the wheel is neglected and is assumed to be in contact with the plane at point $P_{i}$, as according to the initial model assumption nr. 3. In comparison to other wheeled vehicles, the SSWMR has a non-zero lateral velocity. This property stems from the SSWMR's mechanical structure, which necessitates lateral skidding if the vehicle's orientation shifts. As a result, the wheels are only tangent to the path when $\omega = 0$, that is, when the robot travels in a straight line. It is important to consider all wheels together when developing the kinematic model.

Let $\omega _{i}$ and $v_{i}$, with $i = 1,2,3,4$ denote the wheel angular and center linear velocities for front-left, rear-left, front-right, and rear-right wheels, respectively. Thus we have: 
\begin{equation}
\omega _{L} = \omega _{1} = \omega _{2}, \qquad \omega _{R} = \omega _{3} = \omega _{4}
\end{equation}

We can use the previous equation to state the direct kinematics on the plane:

\begin{equation}
\begin{bmatrix}v_{x}\\ v_{y}  \\ \omega_{z} \end{bmatrix} = f\begin{bmatrix}\omega_{l}r\\ \omega_{r}r  \end{bmatrix}
	\label{eq:direct_kin}
\end{equation}
where $v=(v_{x}, v_{y})$ is the vehicle's translational velocity with respect to its local frame, $\omega_{z}$ is its angular velocity  and $r$ is the radius of the wheel.

The instantaneous centers of rotation (ICR) of the left-side, right-side, and robot body are denoted as $ICR_{l}$, $ICR_{r}$ and $ICR_{G}$, respectively, while the mobile robot moves. $ICR_{l}$, $ICR_{r}$ and $ICR_{G}$ are all known to lie on a line parallel to the x-axis~\cite{wong2001general}. We define the x-y coordinates for $ICR_{l}$, $ICR_{r}$ and $ICR_{G}$ as $(x_{ICR_l}, y_{ICR_l})$ , $(x_{ICR_r}, y_{ICR_r})$ , and $(x_{ICR}, y_{ICR})$, respectively. The sides' angular velocity is equal to the velocity of the robot body $\omega _{z}$. We further obtain the following geometrical relations:

\begin{equation}
x_{ICR} = x_{ICR_l} = x_{ICR_r} = -\frac{v_{y}}{\omega_{z}}
	\label{eq:icr_1}
\end{equation}
\begin{equation}
y_{ICR} = \frac{v_{x}}{\omega_{z}}
\end{equation}
\begin{equation}
y_{ICR_l} = \frac{v_{x}-\omega_{l}r}{\omega_{z}}
\end{equation}
\begin{equation}
y_{ICR_r} = \frac{v_{x}-\omega_{r}r}{\omega_{z}}
	\label{eq:icr_4}
\end{equation}

From Equations (\ref{eq:icr_1})–(\ref{eq:icr_4}), the kinematics relation (\ref{eq:direct_kin}) can be represented as:

\begin{equation}
\begin{bmatrix}v_{x}\\ v_{y}  \\ \omega_{z} \end{bmatrix} = J_{\omega}\begin{bmatrix}\omega_{l}r\\ \omega_{r}r  \end{bmatrix}
\end{equation}
where the elements of matrix $J_{\omega}$ are determined by the ICR coordinates on the left and right sides:

\begin{equation}
J_{\omega} = \frac{1}{y_{ICR_l}-y_{ICR_r}}\begin{bmatrix}-y_{ICR_r} & y_{ICR_l} \\ x_{ICR} & -x_{ICR}  \\ -1 & 1 \end{bmatrix}
\end{equation}

Since the SSWMR is symmetrical in our case, we can obtain a symmetrical kinematics model, as seen on the right side of Figure~\ref{fig:skid_steer_kinematic}. As a result, the ICRs are symmetrically distributed on the x-axis, and the matrix $J_{\omega}$ can be written as follows:

\begin{equation}
J_{\omega} = \frac{1}{2y_{ICR_0}}\begin{bmatrix}y_{ICR_0} & y_{ICR_0} \\ 0 & 0  \\ -1 & 1 \end{bmatrix}
\end{equation}
where $y_{ICR_0} = y_{ICR_l} = -y_{ICR_r}$ represents the side ICR values. Considering that for our symmetrical model $v_l=\omega_{l}r$ and  $v_r=\omega_{r}r$, the relations between the angular wheel velocities and the robot velocities are as follows:

\begin{equation}
\begin{cases}v_x = \frac{\omega_{l}r + \omega_{r}r}{2} = \frac{v_l + v_r}{2} \\v_y = 0 \\ \omega_{z} = \frac{-\omega_{l}r + \omega_{r}r}{2y_{ICR_0}} = \frac{-v_l + v_r}{2y_{ICR_0}}\end{cases}
  \label{eq:ctrl_signals}
\end{equation}

Based on equation (\ref{eq:ctrl_signals}), the control signal $u$ can be written as:

\begin{equation}
u = \begin{bmatrix}v_{x}\\ \omega_{z}  \end{bmatrix} = r \begin{bmatrix}\frac{\omega_l + \omega_r}{2} \\ \frac{-\omega_l + \omega_r}{2y_{ICR_0}} \end{bmatrix}
	\label{eq:ctrl_signals_2}
\end{equation}

The last equation shows that the pair of angular velocities $\omega_l$ and $\omega_r$, as well as velocities $v_x$ and $\omega_z$, can technically be viewed as a control kinematic input signal. The accuracy of relation (\ref{eq:ctrl_signals_2}), on the other hand, is heavily reliant on longitudinal slip, and it can only be used if this phenomenon is not dominant. Furthermore, the parameters $r$ and $y_{ICR_0}$ can be calculated experimentally to ensure that the angular robot velocity is accurately estimated in relation to the angular velocities of the wheels.

\section{OctoPath: Architecture, Training and Deployment}
\label{sec:octopath}

\subsection{RNN Encoder-Decoder Architecture}
\label{sec:rnn_encoder_decoder_architecture}

In contrast to traditional neural networks, an RNN's memory cell comprises a time-dependent feedback loop. A recurrent neural network itself can be "unrolled" $\tau_i + \tau_o$ times to produce a loop-free architecture that matches the input length, if we consider an input sequence $[x^{<t-\tau_i>}, ..., x^{<t>}]$ which is time dependant, together with an output sequence $[y^{<t+1>}, ..., y^{<t+\tau_o>}]$. Unrolled networks have $\tau_i + \tau_o + 1$ similar or even identical layers, which means that each layer has the same learned weights.

This architecture is comprised of two models: a stack of several recurrent units for reading the input sequence and encoding it into a fixed-length vector, and a second one for decoding the fixed-length vector and outputting the predicted sequence. The combined models are known as an \textit{RNN Encoder-Decoder}, which is designed specifically for sequence to sequence problems. Given the input sequence $\vec{X}^{<t-\tau_i, t>}$, a basic RNN encoder computes the sequence of hidden states $\left( {{h_1},\;{h_2},\;{h_3},\; \ldots ,\;{h_N}} \right)$:

\begin{equation}\label{eq3}
	{h_t} = \tanh \left( {{{U}_{xh}}{{x}_t} + {{U}_{hh}}{{h}_{t - 1}}} \right)
\end{equation}
\noindent where the two matrices ${{U}_{xh}}$ and ${{{U}}_{hh}}$ are the weight matrix between the input layer and hidden layer, and the weight matrix of recurrent connections  in a given hidden layer, respectively.

The vanishing gradient experienced during training is the major challenge when using simple RNNs. The gradient signal can be multiplied an infinite number of times, up to the number of time steps. As a result, a classical RNN cannot capture long-term dependencies in sequence data. The gradient of the network's output will have a hard time propagating back to affect the weights of the earlier layers if the network is very deep or processes long sequences. The weights of the network will not be successfully modified as a result of gradient vanishing, resulting in very small weight values.

To counter these challenges, in our work we use a set of Long Short-Term Memory (LSTM) networks for both the encoder and the decoder, as shown in Fig.~\ref{fig:neural_network_diagram}. LSTMs solve the vanishing gradient problem by adding three gates that control the input, output, and memory state, as opposed to classical recurrent neural networks.

$\Theta = [W_i, U_i, b_i]$ parametrizes an LSTM network, where $W_i$ embodies the weights of the gates and memory cells multiplied with the input state, $U_i$ represents the weights controlling the network's activations and $b_i$ contains the bias values of the neurons. A network output sequence is defined as a desired ego-vehicle optimal trajectory:

\begin{equation}
	Y^{<t+1, t+\tau_o>} = [y^{<t+1>}, y^{<t+2>}, ..., y^{<t+\tau_o>}]
\end{equation}
\noindent where $y^{<t+1>}$ is a predicted trajectory set-point at time $t+1$. $\tau_i$ and $\tau_o$ are not necessarily equal: $\tau_i \neq \tau_o$.

The LSTM encoder takes the latest octree samples $\vec{X}^{<t-\tau_i, t>}$ as well as the reference trajectory sequence $\vec{Z}_{ref}^{<t-\tau_i, t+\tau_o>}$ for the current time step $t$, and produces an intermediate fixed-size vector ${{c}_t}$ that preserves the temporal correlation of the previous observations. The hidden state of the LSTM encoder ${h_t}$ is calculated using the following equations:

\begin{equation}
 {{{z}}_t} = \sigma ({{{U}}_{xz}}{{{x}}_{t}} + {{{U}}_{hz}}{{{h}}_{t - 1}})
\end{equation}
\begin{equation}
 {{{r}}_t} = \sigma ({{{U}}_{xr}}{{{x}}_{t}} + {{{U}}_{hr}}{{{h}}_{t - 1}}) 
\end{equation}
\begin{equation}
 {{{\tilde h}}_t} = \tanh ({{{U}}_{xh}}{{{x}}_{t}} + {{{U}}_{rh}}({{{r}}_t} \otimes {{{h}}_{t - 1}}))
\end{equation}
\begin{equation}
 {{{h}}_t} = (1 - {{{z}}_t}) \otimes {{{h}}_{t - 1}} + {{{z}}_t} \otimes {{{\tilde h}}_t}
\end{equation}
\noindent where $\sigma$ represents the sigmoid activation function. ${{{z}}_t}$, ${{{r}}_t}$ and ${{{\tilde h}}_t}$ are the update gate, reset gate and candidate activation, respectively. ${{U}}_{xz}$, ${{U}}_{xr}$, ${{U}}_{xh}$, ${{U}}_{hz}$, ${{U}}_{hr}$ and ${{U}}_{rh}$ are the related weight matrices. The notation $\otimes$ represents an element-wise multiplication operator.

The LSTM decoder takes the predicted trajectory sample to produce the subsequent trajectory samples, producing the entire future trajectory $\vec{Y}^{<t+1, t+\tau_o>}$ for the current time step, given the context vector ${{c}_t}$ as input. $\vec{Y}^{<t+1, t+\tau_o>}$ is defined as a sequence variable $Y$ with data instances $[y^{<t+1>}, ..., y^{<t+\tau_o-1>}, x^{<t+\tau_o>}]$ in a specific time interval $[t+1, t+\tau_o]$. Each predicted sequence variable's probability is calculated as follows:

\begin{equation}\label{softmax}
p({{{y}}_t}|{{X}},{{{y}}_{t - 1}}) = g \left ({{{U}}_o}({{E}}{{{y}}_{t - 1}} + {{{U}}_s}{{{s}}_t} + {{{U}}_c}{{{c}}_t})\right )
\end{equation}
\noindent where $g$ is a softmax activation function, as it can be observed in Fig.~\ref{fig:neural_network_diagram}. ${{{s}}_t}$ is the current hidden state of the decoder, ${{{y}}_{t-1}}$ represents the previous target symbol, while ${{E}}$ denotes the embedding matrix.

The earlier target sequence variable ${{{y}}_{t-1}}$ and the context vector $c t$ are also inputs to the decoder, which uses a single unidirectional layer to compute the hidden state $s t$:

\begin{align}\label{al2}
 & {{{z}}'_t} = \sigma ({{{U}}_{yz}}{{E}}{{{y}}_{t - 1}} + {{{U}}_{sz}}{{{s}}_{t - 1}} + {{{C}}_{cz}}{{{c}}_t}) \\
 & {{{r}}'_t} = \sigma ({{{U}}_{yr}}{{E}}{{{y}}_{t - 1}} + {{{U}}_{sr}}{{{s}}_{t - 1}} + {{{C}}_{cr}}{{{c}}_t}) \\
 & {{{\tilde s}}_t} = \tanh ({{{U}}_{ys}}{{E}}{{{y}}_{t - 1}} + {{{U}}_{rs}}({{{r}}'_t} \otimes {{{s}}_{t - 1}}) + {{{C}}_{cs}}{{{c}}_t}) \\
 & {{{s}}_t} = (1 - {{{z}}'_t}) \otimes {{{s}}_{t - 1}} + {{{z}}'_t} \otimes {{{\tilde s}}_t}
\end{align}
\noindent where ${{{z}}'_t}$, ${{{r}}'_t}$ and ${{{\tilde s}}_t}$ are the update gate, reset gate and candidate activation, respectively. ${{U}}_{xx}$ and ${{C}}_{xx}$ are the related weight matrices.

\begin{figure}
	\centering
	\begin{center}
		\includegraphics[scale=1.41]{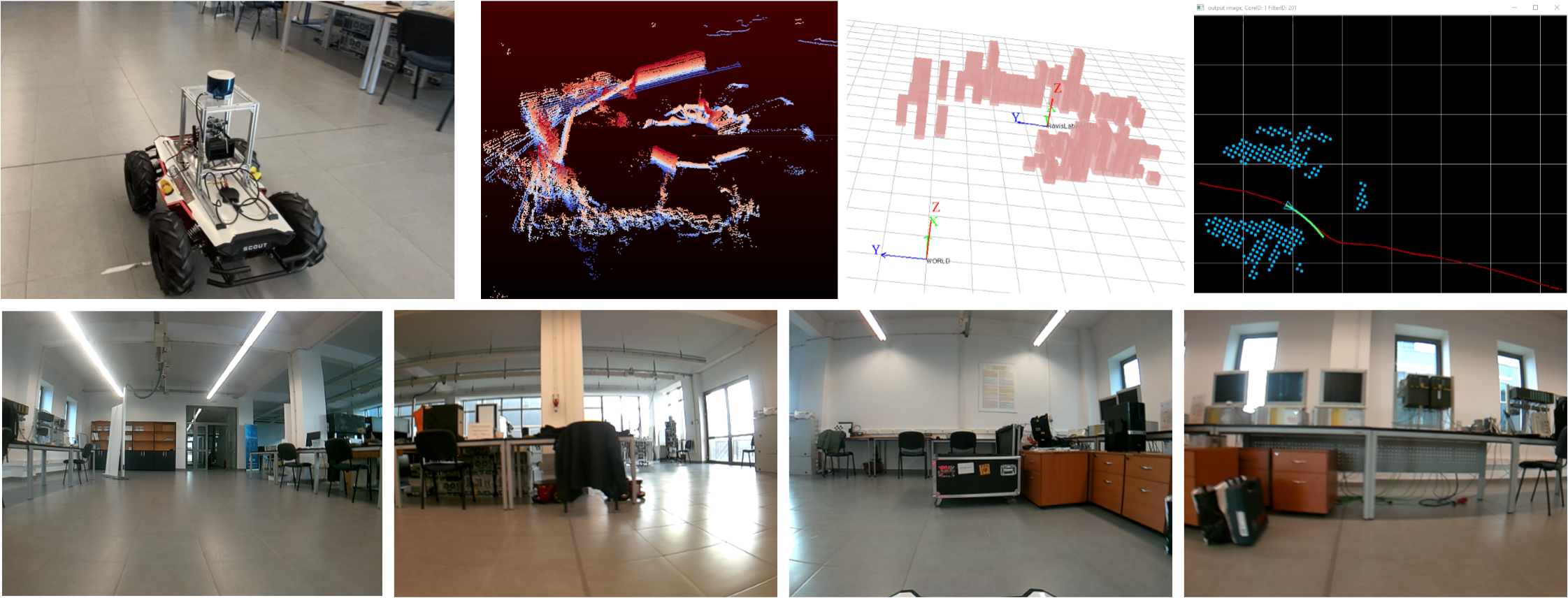}
		\caption{\textbf{Indoor testing setup.} (top) RovisLab AMTU on the reference trajectory, with raw LiDAR data acquired from the top-mounted sensor and the generated 3D and projected 2D octree environment model.  (bottom) The 4 images were acquired with the e-CAM130A quad camera. The order of the images is, given the camera mounting position, front-right-back-left.}
		\label{fig:indoor_testing}
	\end{center}
\end{figure} 

The decoder retains the best sequence candidates in the algorithm when creating the future trajectory sample for each time step. As a result, using the octree input framework, the proposed model would predict the most likely hypotheses of the vehicle trajectory. As analogy to machine translation problems, a point coordinate inside an octree is a character, an octree is a word, and the sequence of input octrees represents an sentence. Our experiments show that an encoder-decoder RNN produces an acceptable trajectory and that its prediction accuracy is improved in comparison to traditional prediction methods.

\subsection{Training Setup}
\label{sec:training}

To train the network, we used data collected with RovisLab AMTU, with the robot being manually driven in the test environment while encountering various static and dynamic obstacles. When acquiring training samples, the point cloud sensory data acquired over Ethernet as UDP packets from a Hesai Pandar LiDAR is used to populate octrees $\vec{X}^{<t-\tau_i, t>}$. Afterwards, together with the global reference trajectory $\vec{z}^{<t-\tau_i, t+\tau_o>}_{ref}$ and with the future position states $\vec{Y}^{<t-\tau_i, t>}$., these are stored as sequence data. For practical reasons, the global reference trajectory is stored at sampling time $t$ over the finite horizon $[t - \tau_i, t + \tau_o]$.

A single training sample, combined from the snapshot of an octree generated from input LiDAR data, maps to $\vec{X}^{<t>}$ and a continuous sequence of octrees is represented further as $\vec{X}^{<t-\tau_i, t>}$. A script implementation is also necessary for processing the sequences from the acquired input data. There is a large number of configurable parameters, such as the sampling time, resolution of the camera, rotations per minute of the LiDAR, and coordinate system for the ego-vehicle data.

We train our method using a self-supervised learning approach, therefore we don't require any manual labeling of the training data. In order to mitigate overfitting, the obstacles were placed differently each time. During training, we have used an 80/10/10 train - validation - test - data split. OctoPath has been trained for 15.000 epochs in a self-supervised fashion using the  stochastic gradient decent method with the Adam optimizer~\cite{kingma2014adam}, with a learning rate of 0.0003. As presented in~\ref{sec:problem_definition}, we express the local trajectory estimation task as a multi-classification problem with N classes (given by the OcTree resolution), thus the loss function to be minimized is given by the negative log-likelihood (NLL) function, as we also have a softmax layer in the last layer of our network. The learning curve of our network (between the training and validation set) can be found on the right side of Fig.~\ref{fig:ablation_study_octree_resolution}.

We used the following hardware configuration to decrease the time required for training: two high-performance graphics cards, namely Nvidia GeForce RTX2080Ti, connected with NVLink, managed by an Intel Core i9-9900K CPU, 64GB RAM, and a 1TB SSD. For the training itself, we have used TensorFlow, because it has seamless integration with the Keras API and is maintained in Python. 

\section{Results}
\label{sec:results}

\subsection{Experimental Setup Overview}
\label{sec:experiments_overview}

\begin{figure}
	\centering
	\begin{center}
		\includegraphics[scale=1.41]{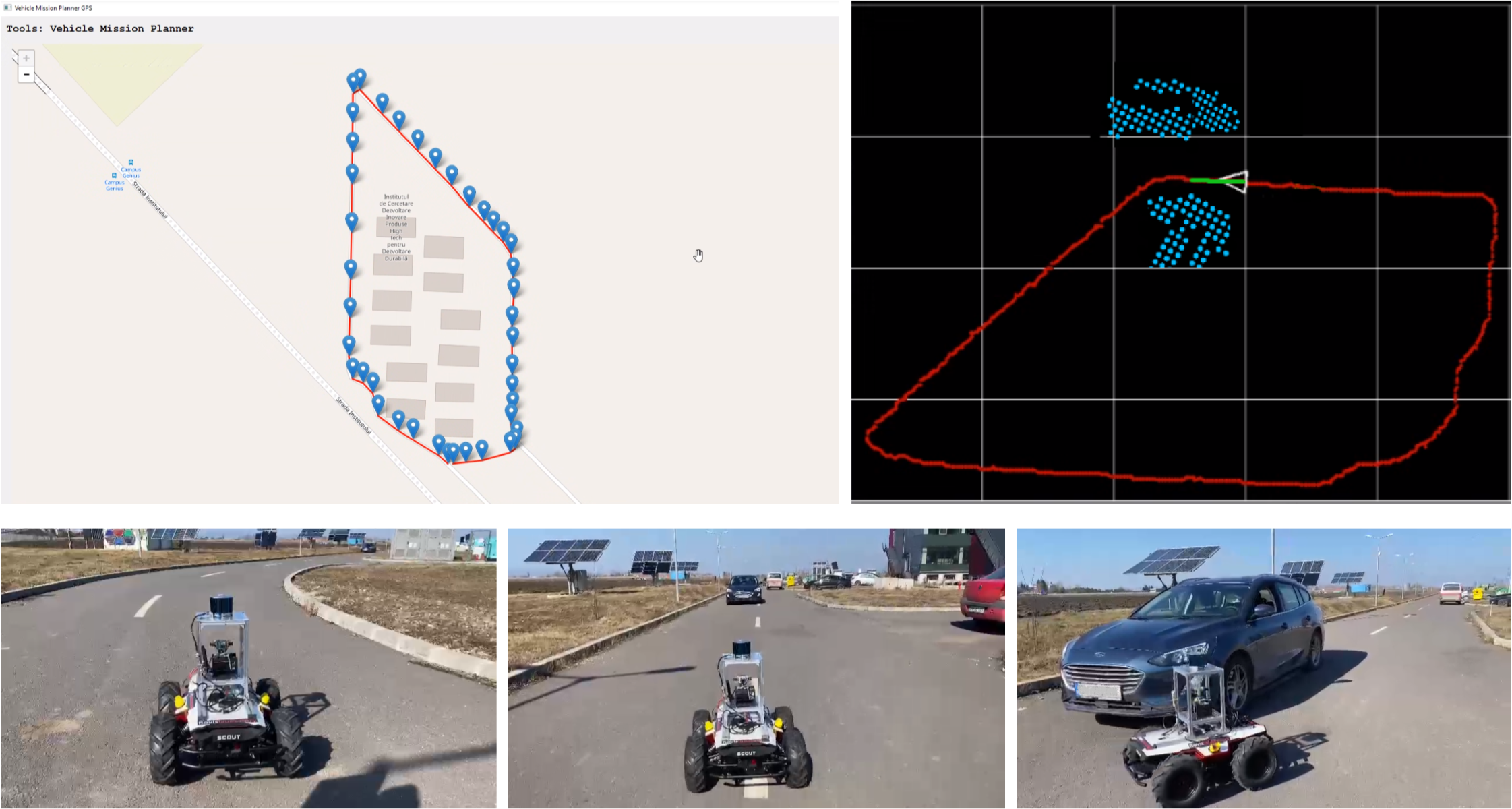}
		\caption{\textbf{Outdoor testing setup.} (top-left) Our vehicle mission planner tool generating the GPS reference route. (top-right) RovisLab AMTU on the reference trajectory with the projected 2D octree environment model. (bottom) On-route behavior with dynamic obstacle avoidance.}
		\label{fig:outdoor_testing}
	\end{center}
\end{figure} 

OctoPath was compared to the baseline hybrid A* algorithm~\cite{dolgov2008practical}, to a regression-based approach~\cite{altche2017lstm}, and to a CNN learning-based approach~\cite{wang2020neural}. We put the OctoPath algorithm to the test in two distinct environments: \textit{I}) in the GridSim simulator~\cite{Trasnea_IRC2019}~\footnotetext{ More information is available at \url{www.rovislab.com/gridsim.html}}(\ref{sec:experiments_I}) and \textit{II}) in a real-world navigation environment, both indoor and outdoor (\ref{sec:experiments_II}), using the RovisLab AMTU robot from  Fig.~\ref{fig:rovis_amtu}. RovisLab AMTU is an AgileX Scout 2.0 platform which acts as a 1:4 scaled car, equipped with a 360\textdegree~Hesai Pandar 40 Lidar, 4x e-130A cameras providing a 360\textdegree~visual perception of the surroundings, a VESC inertial measurement unit, GPS and an NVIDIA AGX Xavier board for data processing and control. The state of the vehicle was measured using wheels odometry and the Inertial Measurement Unit (IMU).

All experiments aimed at solving the trajectory estimation problem illustrated in Fig.~\ref{fig:problem_description}, which was to calculate a trajectory for safely navigating the driving environment without performing the motion control task. To implement motion control, the predicted states were used as input to a model predictive controller, which computed the necessary $v_x$ and $\omega_z$ control signals for the RovisLab AMTU, as detailed in section (\ref{sec:kinematics}). The motion controller's design and implementation are beyond the scope of this paper.

\begin{figure*}
	\centering
	\begin{center}
		\includegraphics[scale=0.975]{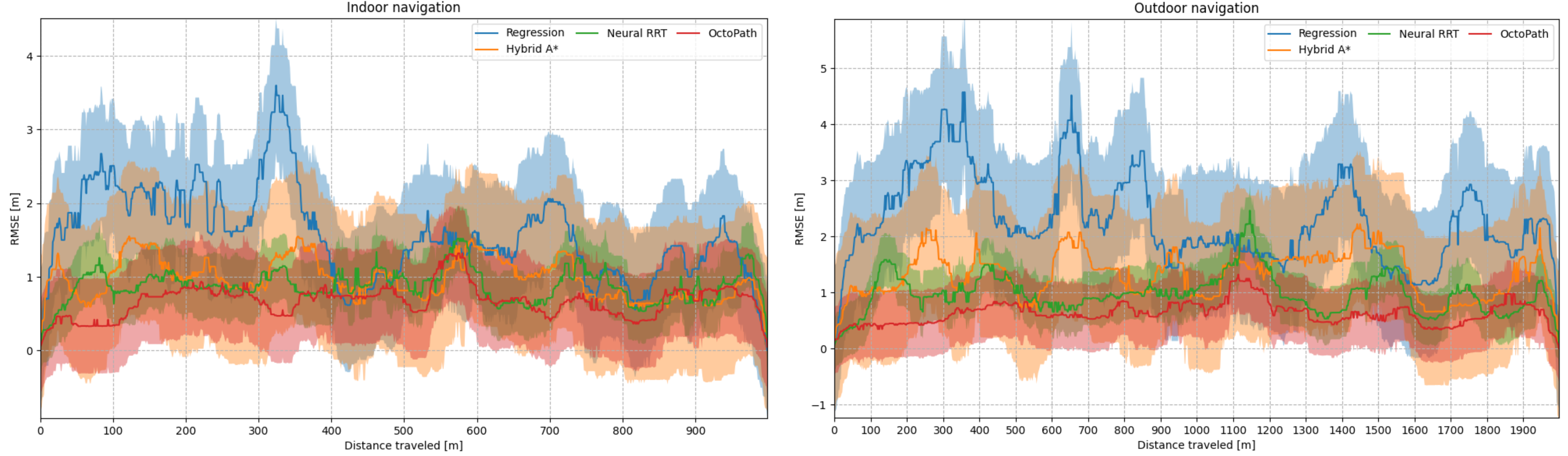}
		\caption{\textbf{Mean (solid line) and standard deviation (shaded region) of the position error, computed as the RMSE from Eq.~\ref{eq:rmse} during indoor and outdoor testing scenarios.}}
        \label{fig:errors}
	\end{center}
\end{figure*}

The hybrid A* algorithm employs a modified state-update rule to apply a variant of the well-known A* algorithm to the vehicle's octree environment model. The search space $(x,y,\theta)$ is discretized, just like in traditional A*, but unlike A*, which only allows visiting cell centers, the hybrid version of the algorithm associates a more continuous state of the car with each grid cell, allowing also trajectory points that are not in the exact center of the octree cell.

In the case of trajectory prediction as a regression problem, the goal is to achieve a direct prediction of continuous future positions without any discretization. Because the average prediction minimizes the regression error, such methods have a bias to output the average of several options, and thus rendering it inaccurate.

The Neural RRT* algorithm, proposed by Wang et. al in~\cite{wang2020neural}, is a novel optimal path planning algorithm based on convolutional neural networks. It used the A* algorithm to generate training data, considering map information as input, and the optimal path as ground truth. Given a new path planning problem, the model can quickly determine the optimal path's probability distribution, which is then used to direct the RRT* planner's sampling operation. The performance of the algorithm varies under different values of the clearance to the obstacles and step size. A wider clearance indicates that the planned route is far from the obstacles, while a smaller clearance indicates that the planned path is closer to them. We have used a fixed step size of 2 and a robot clearance value of 4.

We use the Root Mean Square Error (RMSE) between the predicted and the recorded trajectory in the 2D driving plane:

\begin{equation}
	RMSE = \sqrt{ \frac{1}{\tau_o} \sum^{\tau_o}_{t=1} \left[ (\hat{p}^{<t>}_x - p^{<t>}_x)^2 + (\hat{p}^{<t>}_y - p^{<t>}_y)^2 \right] },
	\label{eq:rmse}
\end{equation}

\noindent where $\hat{p}^{<t>}_x$, $\hat{p}^{<t>}_y$ are the points on the predicted trajectory and $p^{<t>}_x$, $p^{<t>}_y$ are the points on the ground truth trajectory, respectively. We set the prediction horizon $\tau_o = 10$.

The workflow of the experiments is as follows:

\begin{itemize}
\item	collect training data from driving recordings;
\item	generate octrees and format training data as sequences;
\item	train the OctoPath deep network from Fig.~\ref{fig:neural_network_diagram};
\item	evaluate on simulated and real-world driving scenarios.
\end{itemize}

This experimental setup resulted in $15km$ of driving in GridSim, over $1km$ of looped indoor navigation and over $2km$ of outdoor navigation outside of Transilvania University of Brasov's IHTPSD (Institute of High Tech Products for Sustainable Development). The robot navigated indoor and outdoor environments while avoiding static and dynamic obstacles.

\subsection{Experiment I: GridSim Simulation Environment}
\label{sec:experiments_I}

GridSim is a self-driving simulation engine that generates synthetic occupancy grids from simulated sensors using kinematic models, which are then used to produce input octree data. The user interface was integrated into the GridSim environment menu, such that the modes can be switched between replay, record, and training, with each one having access to the five different scenarios. There is a large number of configurable parameters, such as the resolution of the simulator, occupancy grid precision, number of traffic participants, ego vehicle's size, maximum speed, or turning radius.

The first set of experiments compared the four algorithms discussed in~\ref{sec:experiments_overview} over $15km$ of driving in GridSim~\cite{Trasnea_IRC2019}.
The goal is to get from a starting position to a given destination while avoiding collisions and driving at the desired speed. The Z coordinate of all obstacle and free space points is set to zero to adjust the encoder-decoder network's input data to the GridSim environment. The testing scenarios generated using the GridSim simulation environment were not used during the training of the network.

\begin{table}[t]
	\centering
	\begin{tabular}{crccccc}
		\hline
		\textbf{Scenario} & \textbf{Method} & \textbf{$\bar{e}_x[m]$} & $\max(e_x)[m]$ & \textbf{$\bar{e}_y[m]$} & $\max(e_y)[m]$ & \textbf{$RMSE[m]$}\\
		\hline
		 GridSim & Hybrid A* & 1.43 & 3.21 & 2.71 & 4.01 & 2.71 \\
		 simulation & Regression & 3.51 & 7.20 & 4.71 & 8.53 & 5.10 \\
		 & Neural RRT & 1.27 & 3.01 & 2.35 & 2.98 & 2.48 \\
		 & \textbf{Octopath} & \textbf{1.16} & \textbf{2.31} & \textbf{1.72} & \textbf{2.75} & \textbf{2.07} \\
		\hline
		Indoor & Hybrid A*& 1.21 & 4.33 & 1.33 & 3.88 & 1.74 \\
		navigation & Regression & 1.90 & 5.73 & 2.31 & 4.98 & 2.75 \\
		& Neural RRT & 1.01 & 3.29 & 0.98 & 2.16 & 1.44 \\
		& \textbf{Octopath} & \textbf{0.55} & \textbf{1.08} & \textbf{0.44} & \textbf{0.87} & \textbf{0.69} \\
		\hline
		Outdoor & Hybrid A*& 1.35 & 4.67 & 1.44 & 4.44 & 1.98 \\
		navigation & Regression & 2.41 & 8.42 & 2.77 & 8.98 & 3.01 \\
		& Neural RRT & 1.05 & 2.52 & 1.06 & 3.24 & 1.17 \\
		& \textbf{Octopath} & \textbf{0.71} & \textbf{1.46} & \textbf{0.57} & \textbf{1.17} & \textbf{0.88} \\
		\hline
	\end{tabular}
	\caption{Errors between estimated and ground truth trajectories in simulation and real-world navigation testing scenarios.}
	\label{tab:results}
\end{table}

For the various types of roads and traffic environments found in the synthetic testing database, the performance assessment of the benchmarked algorithms is summarized in the top part of the Table~\ref{tab:results}. We illustrate the mean position errors ($\bar{e}_x$, $\bar{e}_y$), as well as the RMSE metric from Eq. (\ref{eq:rmse}).

\subsection{Experiment II: Indoor and Outdoor Navigation}
\label{sec:experiments_II}

The indoor navigation experiment was performed using the RovisLab AMTU SSWMR vehicle from Fig.~\ref{fig:rovis_amtu}, with different indoor navigation tasks. The reference routes which the car had to follow were composed of straight lines, S-curves, circles, and a $75m$ track on the main hallway of Transilvania University of Brasov's Institute for Research, as it can be observed in Fig.~\ref{fig:indoor_testing}.

Clarify the difference between training and testing data. The testing room for the indoor experiment was the same as the one that was used for gathering training data, but the reference routes and the obstacles were placed differently. The main hallway was not used for gathering training data.

The first set of $10$ trials were performed without any obstacles present on the reference routes, while the second $10$ trials set contained static and dynamic obstacles. $54000$ training samples have been collected in the form of LiDAR data and vehicle states, as seen in Figure~\ref{fig:indoor_testing}. The path driven when collecting data was considered as a reference trajectory and was created in a self-supervised manner. 

\begin{figure*}
	\centering
	\begin{center}
		\includegraphics[scale=0.9]{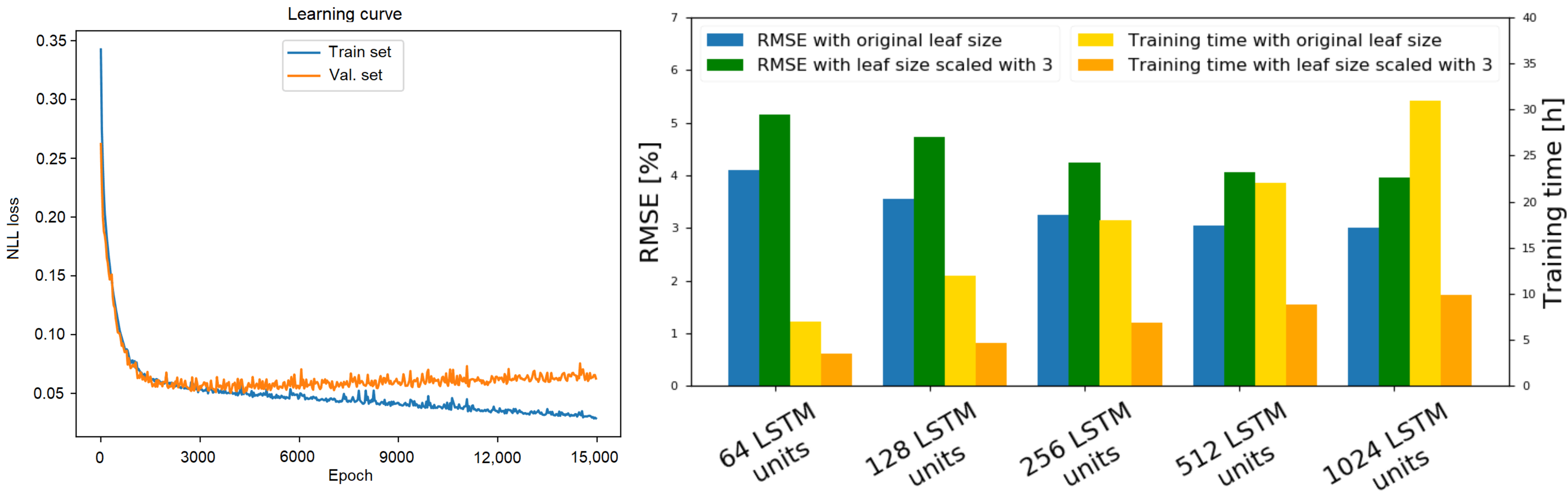}
		\caption{\textbf{Learning curve and ablation of octree resolution and encoder-decoder model parameters. (left side) The evolution of the NLL loss on the training and validation set.}(right side) Performance when training with different numbers of LSTM layers inside both the encoder and the decoder. We see that the RMSE percentages between the estimated and the ground truth driven trajectory decreases with respect to the added number of layers, but gets capped after a certain threshold. Decreasing the resolution causes a small increase in RMSE, but decreases the necessary training time in a significant manner.}
        \label{fig:ablation_study_octree_resolution}
        \vspace{-1.0em}
	\end{center}
\end{figure*}	

The outdoor navigation experiment was performed outside of Transilvania University of Brasov's IHTPSD (Institute of High Tech Products for Sustainable Development), as it can be observed in Figure~\ref{fig:outdoor_testing}. The reference route which the car had to follow was composed of a full loop around the institute and was created using a GPS tool. The route itself is around 500m long, and we ran it 4 times.

The outdoor reference path which was used for training the network was recorded as the driven path when collecting the sensory data. When testing the network, the reference path was generated using our vehicle mission planner tool which can be seen in Fig.~\ref{fig:outdoor_testing}. The static obstacles were mainly parked cars, while the dynamic obstacles were moving cars or people.

The mean and standard deviation of the position error (computed as RMSE) can be viewed in Fig.~\ref{fig:errors}, left side for indoor navigation, and right side for outdoor navigation. The position errors are shown in Table~\ref{tab:results} for all scenarios: simulation, indoor navigation and outdoor navigation. The mean ($\bar{e}_x$, $\bar{e}_y$) and maximum ($\max(e_x)$, $\max(e_y)$) position errors, as well as the RMSE metric from Eq.~\ref{eq:rmse}, are shown. When compared to OctoPath, Neural RRT has the lowest deviations, but from the non-learning approaches, Hybrid A* performs the best, indicating that it is a good candidate for non-learning trajectory estimation.

\subsection{Ablation Study}
\label{sec:ablation}

In this section, we analyse the impact of varying the octree resolution and the network model parameters, as well as the length of the input data sequence from the annotated trajectory dataset. As in the previous section, the RMSE between the estimated and the ground truth driven trajectory is used to quantify the accuracy of our system with respect to the variation of these parameters.

In the first development stage of our algorithm, we have varied the number of LSTM layers inside both the encoder and the decoder, while keeping a fixed number for the output sequence. The experiments testing errors were averaged together to calculate the ablation study from the right side of Fig.~\ref{fig:ablation_study_octree_resolution}. 

The main takeaway from this study is that the performance of the system increases with respect to the number of added layers and neurons. However, the training time also increases at an exponential rate. We have concluded that the optimal structure for the deep network in Fig.~\ref{fig:neural_network_diagram} is composed of $256$ LSTM layers for the encoder and $256$ LSTM layers for the decoder.

\subsection{Deployment of OctoPath on the Nvidia AGX Xavier}
\label{sec:deployment}

\begin{table}[b]
\centering
    {
\begin{tabular}{ccccc}
\hline
\multirow{2}{*}{\begin{tabular}[c]{@{}c@{}}Nvidia AGX Xavier\\ Power Mode\end{tabular}} & \multirow{2}{*}{\begin{tabular}[c]{@{}c@{}}Number of \\ online cores\end{tabular}} & \multirow{2}{*}{\begin{tabular}[c]{@{}c@{}}CPU maximal\\ frequency (MHz)\end{tabular}} & \multirow{2}{*}{\begin{tabular}[c]{@{}c@{}}TensorRT\\ (ms)\end{tabular}} & \multirow{2}{*}{\begin{tabular}[c]{@{}c@{}}Native \\ Tensorflow(ms)\end{tabular}} \\
                                                                                        &                                                                                   &                                                                                        &                                                                          &                                                                                   \\ \hline
MODE\_10W                                                                               & 2                                                                                 & 1200                                                                                   & 41.24                                                                    & 314.66                                                                            \\
MODE\_15W                                                                               & 4                                                                                 & 1200                                                                                   & 29.89                                                                    & 207.12                                                                            \\
MODE\_30W\_4CORE                                                                        & 4                                                                                 & 1780                                                                                   & 21.37                                                                    & 153.86                                                                            \\
MODE\_30W\_6CORE                                                                        & 6                                                                                 & 2100                                                                                   & 17.85                                                                    & 121.38 \\
MODE\_MAXN                                                                              & 8                                                                                 & 2265                                                                                   & \textbf{14.23}                                                                   & \textbf{89.61} \\ \hline
\end{tabular}
}
\caption{Inference time measured on the NVIDIA AGX Xavier with different power mode (nvpmodel) and optimization level settings.}
\label{tab:deployment_agx_xavier}
\end{table}

The results of the network deployment on the Nvidia AGX Xavier mounted on RovisLab AMTU are presented in this section. The board supports three different power modes: 10 watts, 15 watts, and 30 watts, with each mode having several configurations with different CPU and GPU frequencies and a number of available cores. The comparison of OctoPath inference times when using different power modes are shown in Table~\ref{tab:deployment_agx_xavier}. The \textit{TensorRT} column represents the processing time using an optimized model with 4 bytes FP representation (UFF file with the TensorFlow operations replaced by TensorRT plugin nodes), while the \textit{Native Tensorflow} column shows the processing time when using the original TF model (as ProtoBuf file .pb). 

Because of the input OcTree environment model, the network's memory requirements are very low. The native TF ProtoBuf .pb file is slightly less than 10MB in size. OctoPath can be used in embedded systems with real-time performance, as the network will process a frame in 14.23 milliseconds with the Max-N setup. To put it another way, the device is capable of generating 70 paths per second, which is more than enough for most applications. The importance of optimization combined with FP32, as well as the power mode, is emphasized when compared to native TensorFlow solutions.Our robot is more than capable of supplying the current required for the AGX Xavier to operate in MAX-N mode.

\subsection{Discussion}

In our experiments, the hybrid A* algorithm behaved better than the regression approach, mostly because of the structure of the octree environment model input data. This makes A* strictly dependent on the precision of the obstacle representation in the surrounding environment. Besides, the jittering effect of OctoPath may be a side effect of the decoder output's discrete nature. It will, however, provide a reliable ego-vehicle trajectory prediction over a given time horizon.

We intend to use the e-CAM130A synchronized quad cameras in future research to perform a full semantic segmentation on the received point cloud and to extend the validation of our approach to more use-cases. Learning-based approaches, we think, can deliver better results in the long run than conventional methods. This improvement would be achieved by training on more data, which would include a greater number of corner cases.

\section{Conclusions}
\label{sec:conclusions}

OctoPath, our self-supervised method for local trajectory prediction for autonomous vehicles within a finite horizon, is outlined in this paper. A discrete octree-based environment model with configurable resolution provides the input data, which is then fed into a recurrent neural network encoder-decoder. To train the network, we used data collected with a LiDAR sensor mounted on our robot, RovisLab AMTU, a 1:4 scaled vehicle, and we performed simulation and indoor/outdoor navigation experiments. We base our evaluation against a hybrid A* algorithm, a regression-based approach, as well as against a CNN learning-based optimal path planning method, and we conclude that OctoPath is a valid candidate for local trajectory prediction in the autonomous control of mobile robots. Having a configurable environment resolution is an advantage especially on target edge devices, as seen from the deployment on the Nvidia AGX Xavier.

%

\vspace{6pt} 



\authorcontributions{Conceptualization, B.T., C.P. and S.G.; methodology, B.T., C.P. and S.G.; software, B.T., C.G., M.Z., G.M.; validation, B.T., C.G., M.Z. and G.M.; formal analysis, B.T., C.G., C.P. and S.G.; investigation, B.T., C.G., M.Z. and G.M.; resources, C.P. and S.G.; data curation, B.T.; writing---original draft preparation, B.T.; writing, review and editing, C.P. and S.G.; visualization, B.T., C.G. and M.Z.; supervision, G.M., C.P. and S.G.; project administration, C.P. and S.G.; funding acquisition, C.P. and S.G. All authors have read and agreed to the published version of the manuscript.}
\funding{This research received no external funding.}
\institutionalreview{N/A.}
\informedconsent{N/A.}
%
%
\dataavailability{Project website: \url{https://rovislab.com/autonomousvehicles.html}} 
\acknowledgments{This work was supported by the European Union's Horizon 2020 research and innovation programme under grant agreement No 800928, European Processor Initiative
EPI.}
\conflictsofinterest{The authors declare no conflicts of interest.} 
\end{paracol}
\reftitle{References}



\externalbibliography{yes}
\bibliography{references}

\end{document}